\pdfoutput=1

\documentclass[11pt]{article}

\usepackage{NLPAICS2024}

\usepackage{times}
\usepackage{latexsym}
\usepackage{longtable}
\usepackage{rotating}
\usepackage{hyperref}
\usepackage{multirow}
\usepackage{soul}
\usepackage{bbding}
\usepackage{pifont}
\usepackage{wasysym}
\usepackage{amssymb}
\usepackage{enumitem}
\usepackage[T1]{fontenc}

\usepackage[utf8]{inputenc}

\usepackage{microtype}

\usepackage{inconsolata}

\title{Unlocking LLMs: Addressing Scarce Data and Bias Challenges in Mental Health}

\author{Vivek Kumar \and Eirini Ntoutsi\\ 
Research Institute CODE, \\
University of the Bundeswehr, Munich, Germany \\ 
\texttt{\{vivek.kumar,eirini.ntoutsi\}@unibw.de} \And
Pushpraj Singh Rajawat 
\\ Barkatullah University, \\
Bhopal, India\\
\texttt{psrajawatindia@gmail.com} \AND
Giacomo Medda \and Diego Reforgiato Recupero\\
University of Cagliari, Cagliari, Italy \\
\texttt{\{giacomo.media,diego.reforgiato\}@unica.it}\\}

\begin{document} 

\maketitle

\begin{abstract}
Large language models (LLMs) have shown promising capabilities in healthcare analysis but face several challenges like hallucinations, parroting, and bias manifestation. These challenges are exacerbated in complex, sensitive, and low-resource domains. Therefore, in this work we introduce IC-AnnoMI, an expert-annotated motivational interviewing (MI) dataset built upon AnnoMI by generating in-context conversational dialogues leveraging LLMs, particularly ChatGPT. IC-AnnoMI employs targeted prompts accurately engineered through cues and tailored information, taking into account therapy style (empathy, reflection), contextual relevance, and false semantic change. Subsequently, the dialogues are annotated by experts, strictly adhering to the Motivational Interviewing Skills Code (MISC), focusing on both the psychological and linguistic dimensions of MI dialogues. We comprehensively evaluate the IC-AnnoMI dataset and ChatGPT's emotional reasoning ability and understanding of domain intricacies by modeling novel classification tasks employing several classical machine learning and current state-of-the-art transformer approaches. Finally, we discuss the effects of progressive prompting strategies and the impact of augmented data in mitigating the biases manifested in IC-AnnoM. Our contributions provide the MI community with not only a comprehensive dataset but also valuable insights for using LLMs in empathetic text generation for conversational therapy in supervised settings.
\end{abstract}

\section{Introduction} 
Motivational Interviewing (MI) is a client-centered, directive method of conversational counselling that enhances an individual's motivation to achieve behavioural change \citep{miller2012motivational}. MI helps the clients resolve ambivalence and focus on intrinsic motivations by "strengthening client's belief in their capability" or "providing a supportive environment" to make positive changes \cite{moyers2009session,MARTINS2009283,ALPERSTEIN2016393}. MI has gained wide attention from the clinical psychology community due to its proven efficacy in catalyzing significant improvements in health behaviours such as reducing alcohol consumption, smoking cessation, dietary modification, substance abuse, and increasing physical activity \citep{apodaca2014sustain,BARNETT2014498,catley2012motivational,LUNDAHL2013157}. In particular, MI have been very effective in interventions where client adherence and commitment are critical to successful treatment outcomes \citep{hettema2005motivational,tavabi-etal-2021-analysis}. In a nutshell, the core principles of MI, namely, "expressing empathy", "developing discrepancy", "rolling with resistance", and "supporting self-efficacy", are designed to promote a non-confrontational approach that respects client autonomy and facilitates self-directed change \citep{moyers2002motivational}. Since MI is an interactive and time-intensive process, it is accessible to only a small population group, and the reasons account for "\textit{individual's awareness towards mental health}", "\textit{cost of intervention}", "\textit{lifestyle constraints}", and so on. According to World Health Organization report\footnote{\url{https://www.who.int/news-room/fact-sheets/detail/mental-disorders}}, one in every eight people in the world live with a mental disorder and over half (\textbf{54.7\%}) of adults with a mental condition do not have access to effective treatment, summing up over 28 million individuals \cite{world2022world,reinert2021state}. 

Hence, to overcome these challenges and break the barriers in catering to essential and effective treatment, recent research has focused on artificial intelligence (AI) applications. In particular, Large Language Models (LLMs) have been recognised as a potential solution to alleviate the burden on clinicians \citep{10.1093/jamia/ocad258,pmlr-v219-wang23c,Yu2023-sr}. Undoubtedly, LLMs can be instrumental in tackling a wide range of problems directly or by means of assisting roles \citep{doi:10.1073/pnas.2312911120,doi:10.1073/pnas.2300963120}. However, due to its specialised nature, the mental health domain poses unique challenges of complex language understanding that question LLMs efficacy \citep{demszky2023using,bdcc7030124}. Empirical studies have delineated that in such complex domains, LLMs are prone to severe performance concerns like hallucinations \citep{li-etal-2023-halueval,10.1145/3596671.3597650}, stochastic parroting nature \cite{10.1145/3442188.3445922,NEURIPS2023_f26119b4}, and biases \cite{yeh2023evaluating}. 

Therefore, this study aims to bridge this gap by addressing the scarce data and bias challenges in low-resource domains, such as mental health, by generating plausible synthetic data. In this context, we leverage LLMs, particularly ChatGPT and novel prompting strategies, to generate in-context \citep{NEURIPS2020_1457c0d6,chen-etal-2022-improving,dong2022survey} MI dialogues, considering whole therapeutic conversations at once. Furthermore, we develop an evaluation scheme adhering to the Manual for the Motivational Interviewing Skill Code (MISC) \citep{miller2003manual} to assess the quality of generated MI dialogues by comprehensively touching down the psychological and linguistic dimensions. Moreover, we model a novel classification task to identify high- and low-quality MI dialogues. This setting is used to evaluate ChatGPT in terms of domain intricacies understanding, emotional reasoning ability, and biases (contextual, sampling, class imbalance) originated from the experimental dataset. Finally, we discuss the risks of unsupervised employment of LLMs in healthcare, emphasizing the need for collaboration with domain experts and human supervision to ensure responsible LLM implementation across healthcare settings. To put in perspective, our contributions are summarised below: 
\begin{itemize}[leftmargin=12pt, itemsep=2pt, topsep=4pt]
    \item \textbf{Tailored prompting approach}: We propose progressive prompt-based augmentation techniques using LLMs to generate in-context MI dialogue.    
    \item \textbf{Expert annotation}: We develop a rigorous annotation scheme covering psychological and linguistic aspects (e.g., language comprehension, MI structure, false semantics change, contextual reasoning) of generated data grounded on MISC to propose the \textcolor{purple}{\texttt{\textbf{IC-AnnoMI}}} dataset. 
    \item \textbf{Model performance evaluation}: We perform extensive experiments with CML and state-of-the-art (transformer) approaches on the \textcolor{purple}{\texttt{\textbf{IC-AnnoMI}}} dataset to (i) provide a broad set of baselines for the adopted task, (ii) assess the quality of \textcolor{purple}{\texttt{\textbf{IC-AnnoMI}}}, and (iii) discuss potential risks and dangers of unsupervised use of LLMs in sensitive domain. 
    \item \textbf{Reproducibility}: We publicly\footnote{\url{https://github.com/vsrana-ai/IC-AnnoMI}} provide \textcolor{purple}{\texttt{\textbf{IC-AnnoMI}}} and the source code used for our experiments to contribute to the low resource domain and facilitate further research.
\end{itemize}
The rest of the paper is organised as follows. Section~\ref{lit_survey} presents the existing research on LLMs in healthcare. 
Section~\ref{anno_aug} presents the data augmentation, MISC annotation, and the dataset creation. Section~\ref{prob_stat} provides the problem statement and experimental design. Section~\ref{exp_results} outlines our experimental setting and results. Section~\ref{conc_future} addresses the implications of our study and opens up future research directions. Finally, the limitations section discusses the limitations of our work. 

\section{Related work}\label{lit_survey}
In this section, we introduce the works focused on developing public datasets to assist research into psychology and highlight the biases affecting LLMs.

\subsection{Data scarcity in mental health domain}

Domains like psychology and its sub-domains suffer from the scarcity of publicly available resources (datasets) that could be instrumental in mitigating bias in ML approaches and enforcing responsible and ethical AI \citep{10.1145/3395035.3425228}. This problem has gained traction, and researchers have periodically attempted to bridge this gap by developing publicly available datasets. Early efforts in this direction can be credited to \citep{perez-rosas-etal-2016-building}, where they released a dataset annotated with ten counselor behavioural codes of 22,719 counselor utterances extracted from 277 MI sessions. Subsequently, \citep{9746035,fi15030110} released \texttt{AnnoMI}, an expert-annotated GDPR-compliant dataset of 133 high- and low-quality MI sessions. While some of the existing works used \texttt{AnnoMI} to model different tasks \citep{10.1007/978-3-031-37249-0_10} and produce synthetic data \citep{kumar2023data,sdaih23}, some research used it to create further new datasets \citep{hoang-etal-2024-client}. Another study \citep{welivita-pu-2022-curating} released a useful, publicly available dataset of social forums annotated by experts at the therapist statement level with labels adapted from the MITI code \citep{moyers2014motivational}. \citep{yan2022remedi} released ØurResources, a dataset containing 96,965 conversations between doctors and patients, covering 843 types of diseases, 5,228 medical entities, and 3 specialties of medical services across 40 domains. Other notable works in related subdomains contributed with datasets based on textual and conversational settings \citep{sosea2020canceremo,buechel-hahn-2017-readers,bostan-klinger-2018-analysis,bostan-etal-2020-goodnewseveryone,demszky-etal-2020-goemotions,balloccu-etal-2024-ask}.

\subsection{Large language models application and challenge}
LLMs could aid healthcare not only in the workplace but also in enhancing AI systems employed in healthcare. Several studies leveraged LLMs to generate synthetic data to augment the information fed to another model during training~\cite{DBLP:conf/emnlp/LiZL023,DBLP:conf/icdm/Cai0NZ23,DBLP:conf/icdm/WozniakK23,DBLP:conf/eacl/ChowdhuryC24}. A few clinical works explored this methodology and reported satisfying results~\cite{Yuan2023-tg,DBLP:journals/corr/abs-2303-04360,DBLP:conf/emnlp/LiWY23}.
For instance, \cite{DBLP:journals/corr/abs-2303-04360} used LLMs to augment the data for patient-trial matching tasks, while \cite{DBLP:conf/emnlp/LiWY23} proved that LLM-generated data can improve the automatic detection of signs related to Alzheimer's disease from EHRs.
Despite the positive aspects of LLMs, researchers have recently pointed out potential threats associated with using these powerful systems. One of the most concerning factors is the bias in the outcomes of LLMs and AI systems~\cite{DBLP:conf/emnlp/WanPSGCP23,DBLP:conf/kbse/MoralesCC23,DBLP:conf/uemcom/BadyalJC23}, especially when such systems are employed in clinical contexts~\cite{DBLP:journals/csur/SmithKV24,DBLP:journals/ais/GiovanolaT23,10.1007/978-3-031-37249-0_10}. In addition to the prevalent biases such as gender and racial biases, which can lead to misclassifying dosing based on patient ethnicity~\cite{Syn2018-yq} or favoring certain ethnic groups in determining patients-in-need priority scores~\cite{DBLP:journals/ais/GiovanolaT23}, selection and cultural biases are also critical issues~\cite{10.1145/3597307}. These biases can lead to skewed predictions and recommendations, potentially marginalizing minority groups and exacerbating healthcare disparities. 

\section{Data Augmentation, MISC annotation and  dataset creation}\label{anno_aug} 
\begin{figure*}[!htb]
\centering
\includegraphics[width=6in,keepaspectratio]{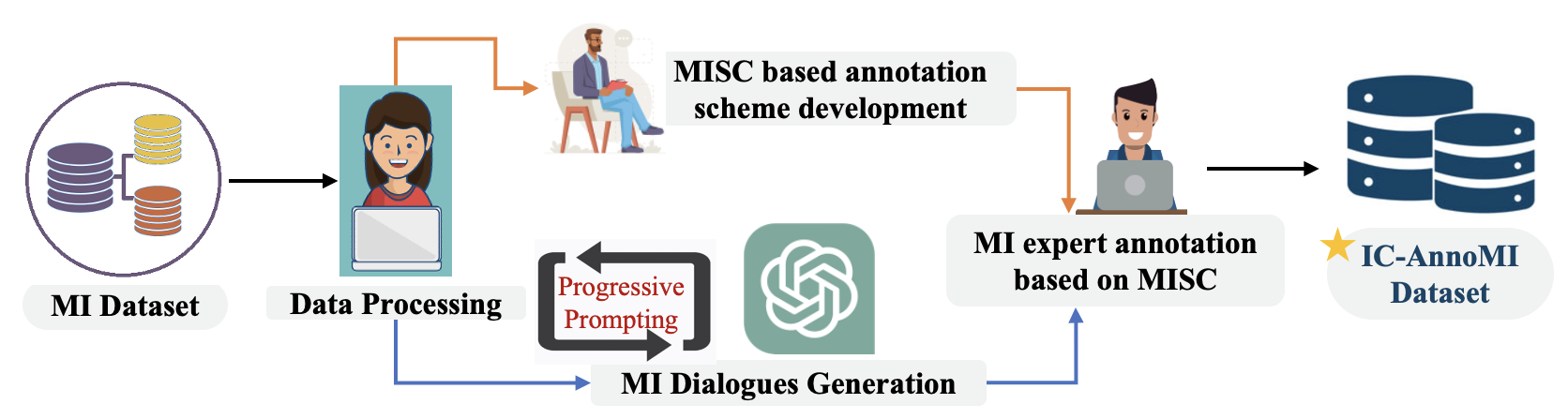}
\caption{Development of the IC-AnnoMI dataset.}
\label{pipeline}
\end{figure*}

In this section, we describe (i) the data augmentation strategy, (ii) how the MISC annotation scheme is developed, and (iii) how the annotation scheme was used to create the dataset. For ease of understanding, Table~\ref{notations_table} outlines the notation used throughout the paper and Figure~\ref{pipeline} depicts the process for the development of the \textcolor{purple}{\texttt{\textbf{IC-AnnoMI}}} dataset.

\begin{table}[!ht]
\centering
\caption{Notations and descriptions/definitions}
\label{notations_table}
\begin{tabular}{p{60pt}p{135pt}}
\hline
\textcolor{purple}{\texttt{\textbf{IC-AnnoMI}}} & The dataset built upon \texttt{AnnoMI} by generating in-context MI dialogues using LLMs progressive prompting. \\
$Client_{utt.}$ & The client utterances in MI dialouges. \\
$Therapist_{utt.}$ & The therapist utterances in MI dialogues. \\
\textcolor{purple}{$MI_{org.}$} & The original MI sessions from \texttt{AnnoMI} dataset. \\
\textcolor{violet}{$MI_{syn.}$} & The generated MI dialogues in \textcolor{purple}{\texttt{\textbf{IC-AnnoMI}}}.\\
\textcolor{red}{$MI_{psych}$} & The parameter representing the psychological aspect of the annotation scheme.\\ 
\textcolor{blue}{$MI_{linguist}$} & The parameter representing the linguistic aspect of the annotation scheme. \\ \hline
\end{tabular}
\end{table}

\subsection{Augmentation} 
\begin{figure}[!htb]
\centering
\includegraphics[width=3in,keepaspectratio]{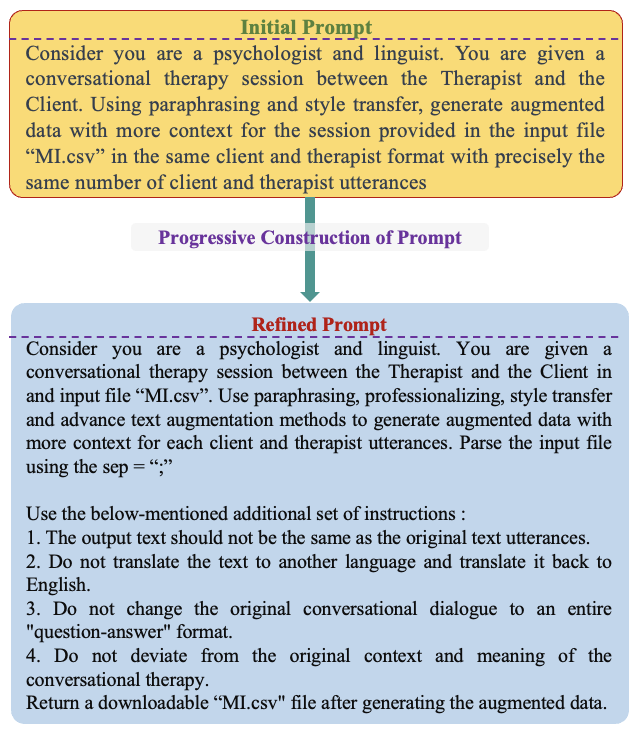}
\caption{Progressive prompt refining.}
\label{propmt}
\end{figure}

The increased quantity of data does not necessarily result in a reliable machine learning (ML) system. Plausible synthetic data can help mitigate inherent biases of experimental datasets such as sampling, contextual, and class imbalance to address the scarce data challenges comprising ML models' reliability. Target augmentation not only provides a better distribution of underrepresented classes but also helps the ML model generalise well. In this research, our primary focus has been context-based augmentation through tailored prompting of ChatGPT variants (4.0 and 3.5 Turbo)\footnote{\url{https://platform.openai.com/docs/models/overview}}. The prompts are engineered through the progressive refinement feedback loop~\citep{song-etal-2023-honeybee,10.1145/3411763.3451760,DBLP:conf/iclr/SuKWSWX0OZS023} until the desired quality and predefined output format are met. In the first step, a prompt template is developed based on MI dialogues' context, plausibility, and quality for required outputs. Then, the generated output is manually evaluated for inconsistencies, and any deviation from the predefined output is used to tune the prompt further progressively. This process continues until the prompt output quality is comparable with \textcolor{purple}{$MI_{org.}$}. For ease of understanding, an example of "initial" and "final" prompt is shown in Figure~\ref{propmt}. Also, to give comparative insights, a sample of \textcolor{purple}{$MI_{org.}$} and \textcolor{purple}{$MI_{syn.}$} is provided in (Appendix~\ref{aapendix_A}).

\subsection{MISC annotation} 
The annotation scheme is developed and executed by an expert from gold-standard institute in psychology by strictly adhering to the MISC 2.1\footnote{\url{https://digitalcommons.montclair.edu/cgi/viewcontent.cgi?article=1026&context=psychology-facpubs}} scheme. The developed annotation scheme is a combination of a two-stage annotation process. The first stage of annotation (\textcolor{red}{$MI_{psych}$}) covers the psychological dimension of the generated MI dialogues. The second stage (\textcolor{blue}{$MI_{linguist}$}) covers the linguistic dimension of MI dialogues. The components of \textcolor{red}{$MI_{psych}$} are further explained as follows.
 
\begin{enumerate}[leftmargin=12pt, itemsep=2pt, topsep=4pt]
    \item \textbf{Empathy}: It is one of the core components of MI and is essential for building rapport and understanding the client's perspective. MI emphasises the therapist's ability to demonstrate empathy through active listening, reflective statements, and genuine curiosity about the client's experiences and feelings \citep{miller2003manual,miller2012motivational}.
    
    \item \textbf{Non-judgmental attitude}: MI encourages therapists to adopt a non-judgmental stance, accepting the client without criticism or a negative attitude. This attitude creates a safe and supportive environment where clients feel comfortable exploring their ambivalence and concerns, which are better captured by a five-point Likert scale. 
    
    \item \textbf{Competence of therapist}: Competence is the therapist's proficiency in applying MI techniques and principles effectively, and it is endorsed by the therapist's experience proven through academic certification and licences \citep{GAUME2009151}.
    
    \item \textbf{Ethical conduct}: In MI practice, ensuring that therapists prioritise the client's well-being, autonomy, and confidentiality is paramount. MI adheres to ethical guidelines established by professional organisations and regulatory bodies such as APA, RCI, etc. These guidelines give the clients autonomy and make sessions more comfortable. Ethical considerations are integral to building trust and maintaining the therapeutic alliance in MI. We follow APA, HIPPA, and other guidelines based on country/region. 
    
    \item \textbf{Reflectiveness}: It involves the therapist's ability to carefully consider and respond to the client's statements, exploring underlying motivations and values. MI encourages therapists to engage in reflective listening and evoke client self-awareness through strategic questioning, which may also include frequent summarisation. Reflective practice enhances the depth and effectiveness of MI interventions, facilitating the meaningful exploration of ambivalence and motivation for change in client sessions.
\end{enumerate}

We have chosen the five-point Likert scale for \textcolor{red}{$MI_{psych}$} annotation because clients can express ambivalent differences in their perceptions, providing more detailed feedback than scales with fewer response options and rather more easily compared with more fine-grained ten-point Likert scale. Indeed, the five-point Likert scale minimises confusion and response errors, facilitating quantitative analysis in terms of mean, standard deviation, and other statistical measures for response summarisation. Compared with ten-pointer scales, converting subjective judgments into five categories enables a clearer alignment with the client's responses and provides sufficient scope to distinguish among different levels of empathy, non-judgmental attitude, competence, ethical conduct, and reflectiveness. \textcolor{red}{$MI_{psych}$} is a numeric value (0-4) averaged over the aforementioned 5 components of \textcolor{red}{$MI_{psych}$} assigned to each \textcolor{violet}{$MI_{syn.}$}. 
The components of {\textcolor{blue}{$MI_{linguist}$}} are binary and can acquire either "Yes" or "No", and these components are briefly mentioned below.
\begin{enumerate}[leftmargin=12pt, itemsep=2pt, topsep=4pt]
    \item \texttt{\textbf{Context}}: It represents the contextual coherence in \textcolor{violet}{$MI_{syn.}$} w.r.t. \textcolor{purple}{$MI_{org.}$}.
    
    \item \texttt{\textbf{Text Enrichment}}: It indicates if \textcolor{violet}{$MI_{syn.}$} is enriched due to style transfer, change in sentence structure, or if more context is added w.r.t. \textcolor{purple}{$MI_{org.}$}.
    
    \item \texttt{\textbf{MI Enhancement}}: It represents if text enrichment and contextual addition has overall enhanced the \textcolor{violet}{$MI_{syn.}$} w.r.t. \textcolor{purple}{$MI_{org.}$}.
    
    \item \texttt{\textbf{$MI_{lang}$}}: It measures if the diction and tone of \textcolor{violet}{$MI_{syn.}$} is preserved and language is refined but avoiding any deviation or false semantic change w.r.t. \textcolor{purple}{$MI_{org.}$}. 
\end{enumerate}
 	
\subsection{Dataset creation} 
For data augmentation, we have used our \texttt{AnnoMI}~\cite{fi15030110}, a publicly available expert-annotated dataset of 133 high- and low-therapeutic counselling dialogues to generate \textcolor{violet}{$MI_{syn.}$}. First, we have filtered out a representative set of \textcolor{purple}{$MI_{org.}$} from \texttt{AnnoMI} considering the high- and low-quality and topic-based distribution of \textcolor{purple}{$MI_{org.}$}, to develop a universal test set for all of our experiments avoiding data contamination. We note that the filtering is done at the MI dialogue level and not at the utterance level to align with our goal of in-context data augmentation, which requires the whole MI dialogue and not the fragments of multiple MI dialogues. This trade-off setup has resulted in 36 \textcolor{purple}{$MI_{org.}$} that constitute the representative test set for our experiments. The remaining 97 \textcolor{purple}{$MI_{org.}$} of \texttt{AnnoMI} constitute the training set and basis of augmentation and MISC annotation. To create \textcolor{purple}{\texttt{\textbf{IC-AnnoMI}}} dataset, the 97 \textcolor{purple}{$MI_{org.}$} of the training set undergo an augmentation process followed by expert annotation using our developed MISC coding scheme. The annotation process overall results in 97 expert-annotated augmented MI dialogues (\textcolor{violet}{$MI_{syn.}$}), containing 4,856 $Therapist_{utt.}$ and 4,792 $Client_{utt.}$ having a mix of high and low-quality MI dialogues. 


\section{Problem Statement and experimental design }\label{prob_stat}
This section presents the problem statement and the research questions we aimed to answer through this research, followed by the dataset description, the applied preprocessing strategies, and the evaluation setup to conduct the experiments. 

\subsection{Problem statement} 
In this work, we primarily focus on classifying high- and low-quality MI dialogues comprised of talk turns between client and therapist at the utterance level, making it a binary classification problem. Therefore, for given $Client_{utt.}$ $\in$ (\textcolor{purple}{$MI_{org.}$}, \textcolor{violet}{$MI_{syn.}$}) and $Therapist_{utt.}$ $\in$ (\textcolor{purple}{$MI_{org.}$}, \textcolor{violet}{$MI_{syn.}$}), the goal is to infer a classification function \textcolor{brown}{$f_{c}$} so that \textcolor{brown}{$f_{c}$} ($Client_{utt.}$, $Therapist_{utt.}$) → \textcolor{blue}{$MI_{quality}$}. Here, \textcolor{blue}{$MI_{quality}$} is the binary class that can only acquire values in $\{0, 1\}$.
The task is designed to evaluate the quality of \textcolor{violet}{$MI_{syn.}$}, the efficacy of LLMs in in-context text generation, and address the below-mentioned research questions.  \\
\textbf{RQ(1)}: How and to what extent do contextual cues and domain-specific prompting strategies help generate real-like MI dialogues? \\
\textbf{RQ(2)}: Can LLMs be used as a potential tool to generate plausible data, considering the whole therapeutic dialogue at once? \\
\textbf{RQ(3)}: How effective is ChatGPT in understanding the complexity of MI dialogues and what are the risks associated with LLMs' employment in sensitive domains?  

\subsection{Dataset preprocessing} 
\begin{figure}[!htb]
\centering
\includegraphics[width=3in,keepaspectratio]{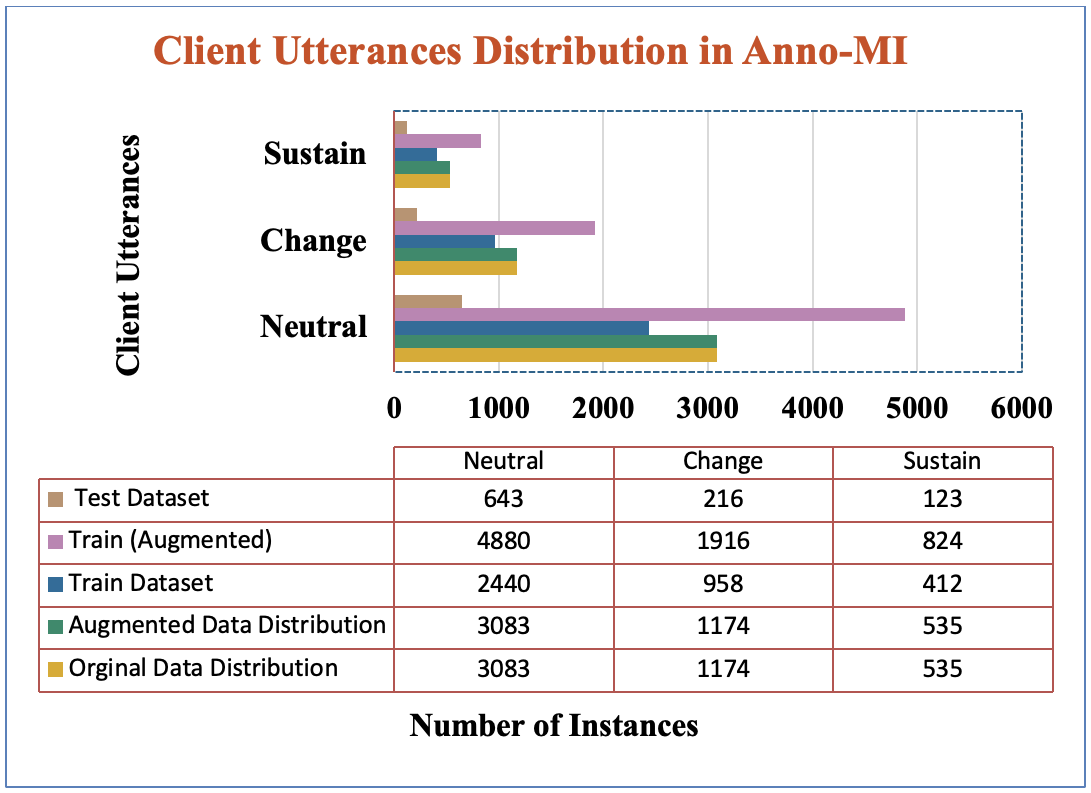}
\caption{The distribution of client utterances in training and test sets of IC-AnnoMI dataset.}
\label{client_distribution}
\end{figure}

\begin{figure}[!htb]
\centering
\includegraphics[width=3in, keepaspectratio]{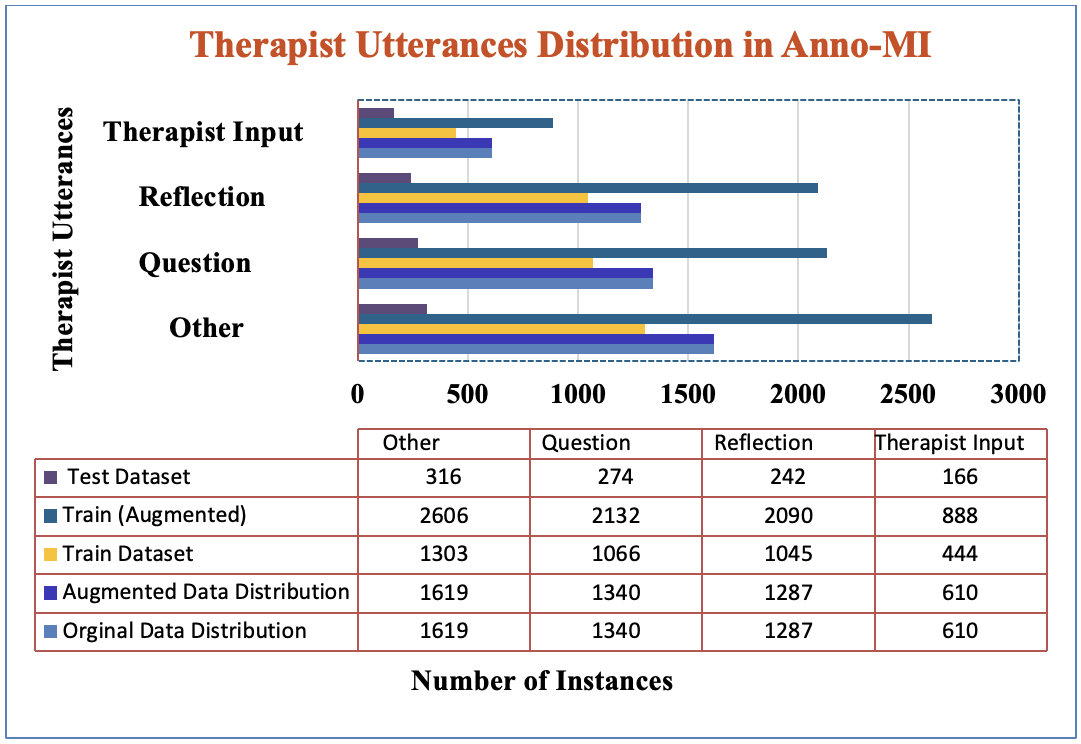}
\caption{The distribution of therapist utterances in training and test sets of IC-AnnoMI dataset.}
\label{therapist_distribution}
\end{figure}

As it can be understood from Figure~\ref{client_distribution} and Figure~\ref{therapist_distribution}, \textcolor{purple}{\texttt{\textbf{IC-AnnoMI}}} has a skewed distribution over target class "high" and "low" quality MI. Also, several MI dialogues have short sentence length in $Client_{utt.}$, $Therapist_{utt.}$, which makes the task more challenging considering the complexity and the small number of MI dialogues. 

Therefore, we have applied tailored preprocessing strategies to avoid semantic loss in $Client_{utt.}$, $Therapist_{utt.}$ and MI dialogue \citep{DBLP:conf/iui/DessiHKRR20,9286431,uysal2014impact,kumar-etal-2023-visu}. The preprocessing steps include lowercasing the text for uniform representation (e.g., Psychology and psychology have a common representation → psychology). We have removed punctuation, whitespaces, newlines, and extra space,s but retained stopwords. This design choice relies on the fact that MI dialogues in \textcolor{purple}{\texttt{\textbf{IC-AnnoMI}}} have several short $Client_{utt.}$, $Therapist_{utt.}$, up to 3 tokens length. Thus, removing stopwords (e.g., not) might change the whole course of the conversation, contributing to misclassification errors. We have also removed multilingual symbols, special characters, elements not part of the standard English language, and expanded contractions such as \textit{it's} --> \textit{it is}.

\begin{table*}
\centering 
\resizebox{\textwidth}{!}{
\begin{tabular}{c|l|c|c|c|c|c|c|c|c|c|c}
\hline
\multirow{2}{*}{\rotatebox[origin=c]{90}{\textbf{Emb.}}} & \multirow{2}{*}{\textbf{Model}} & \multicolumn{2}{c}{\textbf{Acc.}} & \multicolumn{2}{c}{\textbf{Bal. Acc}} & \multicolumn{2}{c}{\textbf{Precision}} & \multicolumn{2}{c}{\textbf{Recall}} & \multicolumn{2}{c}{\textbf{F1-Macro}} \\
& & \textbf{N-Aug.} & \textbf{Aug.} & \textbf{N-Aug.} & \textbf{Aug.} & \textbf{N-Aug.} & \textbf{Aug.} & \textbf{N-Aug.} & \textbf{Aug.} & \textbf{N-Aug.} & \textbf{Aug.} \\
\hline
\multirow{2}{*}{\rotatebox[origin=c]{90}{NA}} & Naive Bayes & .80 & .83 & .49 & .50 & .83 & .83 & .80 & .83 & .81 & .83 \\
& Random Forest & .89 & .89 & .51 & .50 & .84 & .84 & .89 & .90 & .86 & .86 \\
\hline
& & & & & & & & & & & \\
\multirow{1}{*}[2mm]{\rotatebox[origin=c]{90}{Static}} & BiLSTM (word2vec) & .87 & .87 & .50 & .50 & .83 & .83 & .87 & .87 & .85 & .85 \\
& & & & & & & & & & & \\
\hline
\multirow{6}{*}{\rotatebox[origin=c]{90}{Contextual}} & $BERT_{base}$ & .89 & .90 & .54 & .56 & .86 & .87 & .89 & .90 & .87 & .88 \\
& BART & .87 & .89 & .54 & .57 & .86 & .86 & .86 & .89 & .87 & .87 \\
& DistilBERT & .89 & .89 & .55 & \textbf{.59} & .86 & .87 & .89 & .89 & .87 & .88 \\
& AlBERT & .89 & .90 & .52 & .55 & .85 & .87 & .89 & .90 & .87 & .88 \\
& RoBERTa & .88 & .90 & .54 & .57 & .86 & .86 & .88 & .90 & .87 & .87 \\
& XLNet & .88 & .88 & .54 & .57 & .85 & .86 & .88 & .88 & .86 & .87 \\ \hline
\end{tabular}
}
\caption{The results of CML and DL approaches with the non-augmented (\texttt{N-Aug}) and augmented (\texttt{Aug}) dataset.}
\label{result_aug_non_aug_data}
\end{table*}

\subsection{Experiments} 
We have employed various classification models for our experiments, including CML and transformer-based models, to provide a baseline and optimal experimental setup for such task in therapeutic settings. In CML, we have used Support Vector Machine, Naive Bayes, and Random Forest. In deep learning (DL), we used a BiLSTM-based deep neural network architecture with pre-trained word embeddings\footnote{\url{https://code.google.com/archive/p/word2vec/}} for feature representation. For transformer-based models, we have employed $BERT_{base}$ \citep{devlin-etal-2019-bert}, and some of its variants, such as \textit{DistilBERT} \citep{sanh2019distilbert}, \textit{RoBERTa} \citep{liu2019roberta}, \textit{AlBERT} \citep{lan2019albert}, \textit{BART} \citep{lewis-etal-2020-bart}, and  \textit{XLnet} \citep{NEURIPS2019_dc6a7e65}, using python libraries such as \textit{Keras\footnote{\url{https://keras.io/}}}, Tensorflow\footnote{\url{https://t fhub.dev/google/collections/bert}}, and ML platforms like Hugging Face\footnote{\url{https://huggingface.co/docs/transformers/index}}. The metrics used to evaluate the performance of implemented ML models are accuracy, balanced accuracy, precision, recall and F1-Score and the formulas  are provided in (Appendix~\ref{aapendix_B}). The training, validation and test distribution for all the experiments are ~$63\%$, ~$10\%$, and ~$27\%$ respectively, and the computational resource used to conduct the experiments is mentioned in (Appendix~\ref{aapendix_C}). 
\section{Result and discussion}\label{exp_results} 
In this section, we provide insights from our results and in-depth analyses based on our experimental outcomes. The classification results of the implemented ML models with the non-augmented and augmented \textcolor{purple}{\texttt{\textbf{IC-AnnoMI}}} datasets are summed up in Table~\ref{result_aug_non_aug_data}.

Note that the applied augmentation method is not centered on reducing the class imbalance in the experimental dataset by targeting the minority class, which is \textbf{low-quality} MI in our case, but on preserving the context of each dialogue. Therefore, this augmentation is not expected to contribute significantly to applied ML models' performance, but to have more of an impact on increasing the sample size of the training set. The main experimental observations are as follows: 

\begin{itemize}
\item \textbf{Performance of CML models}: The CML models trained on 2,456 features have shown to be ineffective in accurately identifying the high- and low-quality MI, with a high misclassification rate towards the minority class, as evident from the confusion matrices shown in Figure~\ref{cml_conf} as expected. The reason is that the features selected in the bag-of-words approach are given weightage based on occurrence frequency, which in complex domains do not sufficiently capture the context of the entire MI dialogue. 

\item \textbf{Performance of DL (BiLSTM) model}: The DL model has also not shown much improvement over CML models due to the fact that the text length of utterances is small, the dataset is very imbalanced, and the number of training MI samples are far too less for a DNN based model to learn and generalise well for such complex domain. 

\item \textbf{Performance of $Bert_{base.}$ and its variants}: This is where the advantage of augmentation reflects. All the language models (LMs), namely $Bert_{base.}$, BART, DistilBERT, ALBERT, RoBERTa, and XLNet, have shown improvement in the performance. In particular, the increase in balanced accuracy is indicative of better generalisation and mitigation of inherent bias in \textcolor{purple}{\texttt{\textbf{IC-AnnoMI}}}. Although all the models have comparable scores in terms of balanced accuracy, DistilBERT has scored the highest, which is \textbf{0.59}. A comparative insight through confusion matrices is presented in Figure~\ref{bert_cof}. The observed improved performance in employed LMs verifies that the quality of \textcolor{violet}{$MI_{syn.}$} is in line with \textcolor{purple}{$MI_{org.}$}. 

\begin{figure*}[!htb]
\centering
\includegraphics[width=5in,keepaspectratio]{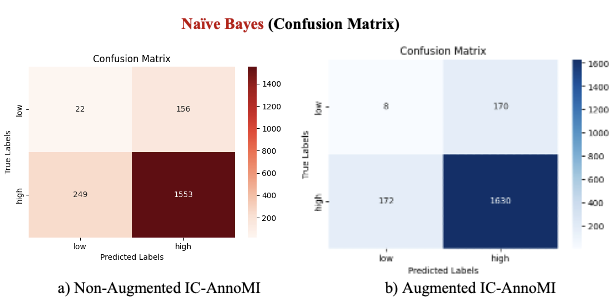}
\caption{The confusion matrix of CML approaches for non-augmented and augmented experimental datasets.}
\label{bert_cof}
\end{figure*}

\begin{figure*}[!htb]
\centering
\includegraphics[width=5in, keepaspectratio]{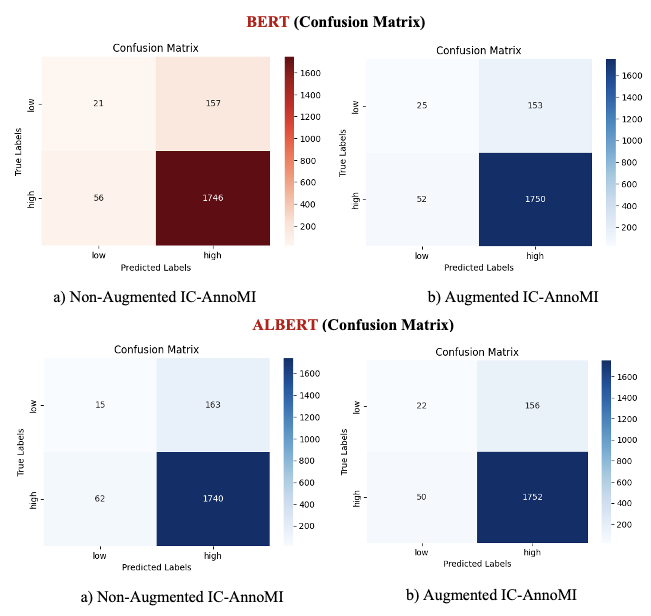}
\caption{The confusion matrix of BERT model-based approaches for non-augmented and augmented experimental datasets.}
\label{cml_conf}
\end{figure*}

\item \textbf{Performance based on expert evaluation}: The statistics of expert annotated components of \textcolor{red}{$MI_{psych}$} and \textcolor{blue}{$MI_{linguist}$} of \textcolor{violet}{$MI_{syn.}$} are also in agreement with the above performance, which strengthens our results. For instance, {$MI_{psych}$} has received an average score of \textbf{3.31} for the 97 \textcolor{violet}{$MI_{syn.}$} averaged over its five attributes and then averaged over 97 \textcolor{violet}{$MI_{syn.}$}. Also, for the \textcolor{blue}{$MI_{linguist}$} aspect of 97 \textcolor{violet}{$MI_{syn.}$}, \textbf{95.88\%} have preserved the \texttt{\textbf{Context}}, \textbf{83.51\%} have contributed to \texttt{\textbf{Text Enrichment}}, \texttt{\textbf{MI Enhancement}} is observed in \textbf{88,66\%} and overall \texttt{\textbf{$MI_{lang}$}} is \textbf{88,66\%}. 

\item \textbf{Answer to the research questions}: These high scores of {$MI_{psych}$} and \textcolor{blue}{$MI_{linguist}$} are answers to research questions \textbf{RQ(1), RQ(2)} and \textbf{RQ(3)}. The experimental outcomes indicate that contextual cues and domain-specific prompting strategies can help generate dialogues qualitatively close to \textcolor{purple}{$MI_{org.}$}. LLMs, in our case, ChatGPT, are considerably successful in understanding the fine-grain intricacies of MI and comprehending the flow, context, and nuances of therapeutic settings. However, we also observed inconsistencies in this experimental process at the stage of prompt designing, when hallucinations, absurd text generation, and stochastic parroting happened until they were humanly identified and eliminated through rigorous prompt refining. 
\end{itemize}

\section{Conclusion and future work}\label{conc_future} 
This paper explores LLMs' capabilities, particularly ChatGPT, for data augmentation in mental health and therapeutic counselling scenarios. Through this research, we seek to study the operability of LLMs in solving the data scarcity issue in therapeutic counselling and verify that biases are not reinforced when models are trained on LLM-generated synthetic data. To this end, we employed a progressive prompt technique to generate in-context plausible MI dialogues and further expert annotated them by developing a comprehensive MISC coding scheme considering MI sessions' psychological and linguistic aspects. To evaluate the quality of generated MI dialogues and to understand to what extent the generated dataset is relevant to the annotation scheme, we employed several CML and transformer-based models to establish a baseline for the classification task of MI dialogues' quality at the utterance level. Our results highlight the efficacy of the augmentation and annotation scheme, given that the augmented dataset led to improvements in classification and mitigation of inherent biases. The findings demonstrate that the data generated through this rigorous quality control process is both plausible and substantially beneficial in enabling ML techniques to address the targeted biases, thereby supporting the use of LLMs for supervised, task-specific applications in sensitive domains like mental health. However, despite the favorable outcomes, risks and concerns are associated with the unsupervised application of LLMs in sensitive domains, and it is thus advised to use them with humans in the loop to promote responsible and ethical AI uses. The future research direction is set to explore other LLMs such as Mistral \citep{karamcheti2021mistral}, Falcon \citep{almazrouei2023falcon}, LLama \citep{touvron2023llama}, etc., to understand their reliability in mental health domain and plausible data generation. We also aim to tackle MI dialogue-based classification instead of utterance-based and integrate domain knowledge \citep{9866735} in classification systems generated by LLMs to tackle domain adaptation problems.  

\section*{Limitations}\label{limitations} 
While our work provides a holistic novel annotation scheme adhering to MISC to create and annotate synthetic MI dialogues, covering both the psychological and linguistic dimensions, it has some limitations and room for improvement. The main limitation can be considered as the low number of MI sessions, which may lead to sub-optimal performance and biases in ML approaches. Another limitation is the computational resource that may have hampered the LMs from being used at their full potential. So we consider using larger resources to avoid this limitation. In this work our focus is in-context dialogue MI generation at the session level that necessarily reduces the class imbalance. Therefore, we aim to generate MI dialogues targeting underrepresented classes leveraging different LLMs to be in more contextual diversity. 

\section*{Ethics statement}
\subsection{Expert Annotation}
To maintain the integrity and quality of the data, qualified experts affiliated with the gold-standard organization in psychology have performed the annotations. The experts have significant experience and training in MI to ensure therapy’s nuances and ethical considerations are appropriately enforced in the annotation process. The expert is also bound by confidentiality agreements to safeguard the privacy of the individuals in the MI recordings and transcripts.
\subsection{Ethical Concerns}
We acknowledge that our work has strictly followed the norms and protocols of ethical considerations throughout the research process.  We also enforce adherence to ethical standards and guidelines for researchers who want to use our data to ensure ethical and responsible use of the resource.

\section*{Acknowledgements}
This research work is funded by the European Union Horizon Europe Project STELAR, Grant Agreement ID: 101070122.



\bibliography{biblio}
\bibliographystyle{acl_natbib}

\appendix

\vspace{2in}

\section{Appendix}
\label{aapendix_A}
\begin{figure}[!htb]
\centering
\includegraphics[width=3in,keepaspectratio]{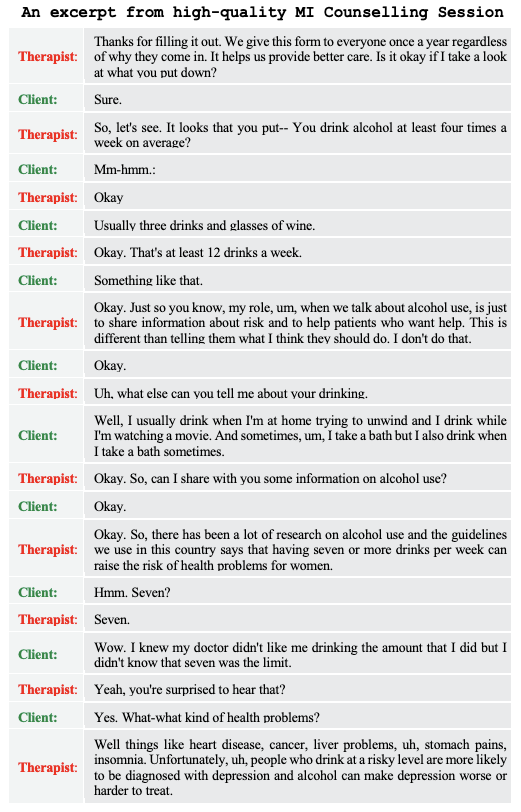}
\caption{The distribution of client utterances in training and test set of IC-AnnoMI dataset.}
\end{figure}

\begin{figure}[!htb]
\centering
\includegraphics[width=3in, keepaspectratio]{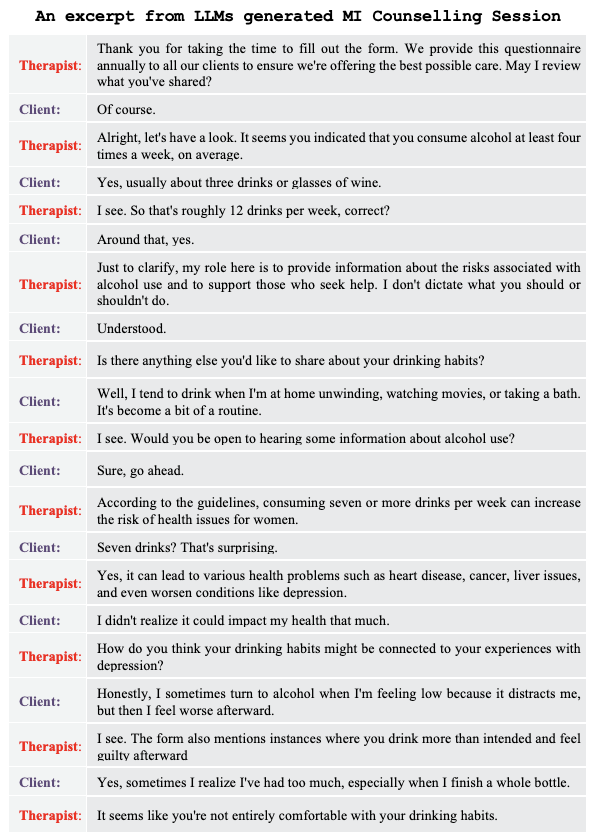}
\caption{The distribution of therapist utterances in training and test set of IC-AnnoMI dataset.}
\end{figure}

\section{Appendix}
\label{aapendix_B}
\begin{equation}Accuracy=\frac{TP+TN}{TP+TN+FP+FN}\end{equation}
\begin{equation}Bal. Acc.= \frac{TP(TN+FP)+TN(TP+FN)}{2\times(TP+TN+FP+FN)}\end{equation}
\begin{equation}Precision=\frac{TP}{TP+FP}\end{equation}
\begin{equation}Recall= \frac{TP}{TP+FN}\end{equation}
\begin{equation}F1=2 \times \frac{Precision \times Recall }{Precision + Recall}\end{equation}
where TP stands for true positive, TN for true negative, FP for false positive, and FN for false negative.

\vspace{2in}
\section{Appendix}
\label{aapendix_C}
\begin{table}
\label{serverspecification}
\setlength{\tabcolsep}{3pt}
\begin{tabular}{p{65pt}p{135pt}}
\hline
\textbf{Item} & \textbf{Specification}\\
\hline
CPU & Intel Core i3-7100 (-HT-MCP-) CPU @ 3.90 GHz \\
GPU	& NVIDIA GP102 [TITAN X], 12 GB memory \\
Graphic Driver & NVIDIA graphic driver version 440.33.01 \\
CUDA & Version 10.2 \\
OS & Ubuntu (17.10)\\
Python & Version 3.6.6\\
\hline
\end{tabular}
\caption{Server specifications.}
\end{table}

\end{document}